# Improved Estimation in Time Varying Models

**Philip Bachman** PHIL.BACHMAN@GMAIL.COM
**Doina Precup** DPRECUP@CS.MCGILL.CA
School of Computer Science, McGill University, Montreal, Canada

## Abstract

Locally adapted parameterizations of a model (such as locally weighted regression) are expressive but often suffer from high variance. We describe an approach for reducing this variance, based on the idea of estimating simultaneously a transformed space for the model and locally adapted parameterizations expressed in the new space. We present a new problem formulation that captures this idea and illustrate it in the important context of time varying models. We develop an algorithm for learning a set of bases for approximating a time varying sparse network; each learned basis constitutes an archetypal sparse network structure. We also provide an extension for learning task-specific bases. We present empirical results on synthetic data sets, as well as on a BCI EEG classification task.

## 1. Introduction

Locally adapted parameterizations can produce flexible representations from relatively rigid components; locally weighted regression serves as a canonical example of this approach. Such models reduce bias but increase variance, due to reduced effective sample sizes used for each estimation. We tackle this problem using a natural machine learning idea: using a transformed (more restricted or simpler) space in which to find local parameterizations.

A common approach to improving model efficacy in machine learning is to first transform the *data* into an alternate representation prior to model estimation, ideally in a way that amplifies useful information while attenuating noise. Algorithms exemplifying this approach include: PCA, ICA (Hyvärinen & Oja, 2000), nonlinear-dimension reduction, e.g. (Tenenbaum et al., 2000), and dimension reduction for regression (Fukumizu et al., 2004; Cook & Forzani, 2009).

 

Another line of work considers transformations of the *model* used to describe the data, either by reducing the number of degrees of freedom, or by seeking a model form amenable to more powerful estimation procedures. Examples of the first approach include DiscLDA (Lacoste-Julien et al., 2008) and supervised dimensionality reduction using Bayesian mixture models (Sajama & Orlitsky, 2005), which seek useful linear reductions of the parameters of a generative model. The second approach includes the application of spectral methods to learning transformed representations of HMMs (Siddiqi et al., 2010) and PSRs (Boots & Gordon, 2011).

In this paper, we provide a different lens through which to view model transformations. In Sec. 2, we present a general formulation of the problem of estimating useful transformations of model parameters, which encompasses several of the previously mentioned methods for both data and model transformation. Our problem depends on the simultaneous estimation of a transformation of the parameter space of a model and of the parameters within the transformed space. We formulate the problem primarily for use with multiply parameterized models (such as locally weighted linear regression or mixture models), which distinguishes it from the spectral methods for HMM and PSR learning, which seek *single* transformed parameterizations of a given model. We illustrate our problem formulation in the context of familiar models (locally weighted regression and Gaussian mixtures) in Sec. 3. In Sec. 4 we present a novel algorithm for modeling time varying sparse network structures underlying sequential observations. In Sec. 5 and 6, we use synthetic data and data drawn from real-world BCI EEG experiments to showcase our algorithm.

## 2. A General Problem Formulation

The problem investigated in this paper arises as a generalization of the following optimization:

$$B^* = \arg\min_B \left[\ell(f, X, B)\right] \qquad (1)$$

where the loss $\ell$ measures the "goodness" of fit of the model $f$ to the data $X = \{(x_1, y_1), ..., (x_m, y_m)\}$ given a *set of parameterizations* $B = \{\beta_1, ..., \beta_{m'}\}$ of $f$, and an optimal set of parameterizations $B^*$ is sought.

The idea of using multiple model parameterizations is not often explored in machine learning. As motivation for this view, we begin by expressing standard linear regression in the form of (1). In this case, $f$ measures the residuals produced by a parameter vector:

$$f((x, y), \beta) = \beta^T x - y$$

For a set of parameter vectors $\beta_i$, $\ell$ is proportional to the log-likelihood of observing the residuals assuming they are normally distributed with variance $\sigma^2$:

$$\ell(f, X, B) = \frac{1}{\sigma^2} \sum_{i=1}^{m'} \sum_{j=1}^{m} f((x_j, y_j), \beta_i)^2 \qquad (2)$$

We usually think of the loss in this case as having $m' = 1$. However, note that considering $m' > 1$ does not modify the solution, as loss is measured equally over all $(x_j, y_j)$, which implies that $\beta_i = \beta_j, \forall \beta_i, \beta_j \in B^*$ (i.e., there still is, in effect, one optimal parameter vector).

Using this view, we can transform standard linear regression into kernel weighted linear regression as follows:

$$f((x, y), \beta) = \beta^T x - y$$
$$\ell(f, X, B) = \frac{1}{\sigma^2} \sum_{i=1}^{m'} \sum_{j=1}^{m} k(\beta_i^x, x_j) f((x_j, y_j), \beta_i^w)^2 \qquad (3)$$

where the kernel weighting function $k(x, x')$ measures similarity between locations in the input space, and each $\beta_i$ consists of two components: $\beta_i = (\beta_i^w, \beta_i^x)$. The localization component $\beta_i^x$ associates $\beta_i$ with a location in the input space and the coefficient component $\beta_i^w$ associates $\beta_i$ with a set of regression coefficients.

Introducing the kernel $k$ allows the $\beta_i$ in (3) to have local rather than global effect, which leads to different parameterizations at each location in observation space. However, there is no need to optimize jointly over $B^*$, as there is no constraint linking different elements in a parameter set. Allowing multiple local parameterizations of $f$ is useful for increasing the power of simple models; estimation of time varying covariance matrices in financial modeling and estimation of time varying auto-regressions in econometrics are two well-studied examples of this idea.

While locally weighted regression is typically thought of as a "non-parametric" method, in the context of our work it is more fruitfully viewed as an approach based on multiple parameterization, in which the implied infinite set $B^*$ can be queried "lazily" for specific parameter locations $\beta_i^x$, rather than computed monolithically.

To illustrate a problem in the form of (1) in which the elements of $B^*$ are not independent, for $\beta_i = (\beta_i^\mu, \beta_i^\Sigma, \beta_i^\pi)$, consider the following optimization:

$$f((x, y), \beta) = \beta^\pi p(x|\beta^\mu, \beta^\Sigma)$$
$$\ell(f, X, B) = -\log\left(\prod_{j=1}^{m}\left[\sum_{i=1}^{m'} f((x_j, y_j), \beta_i)\right]\right), \qquad (4)$$

in which $0 \leq \beta_i^\pi \leq 1, \forall \beta_i$ and $p(x|\beta^\mu, \beta^\Sigma)$ is the probability of observing $x$ given a Gaussian distribution with mean $\beta^\mu$ and covariance $\beta^\Sigma$. Minimizing (4) corresponds to estimating a Gaussian mixture model for the data $X = \{x_1, ..., x_m\}$. Interdependence among the $\beta_i \in B^*$ is induced by the negative log-likelihood loss, together with a constraint on the set of mixture weights: $\sum_i \beta_i^\pi = 1$.

Note that in the last two examples, the estimation of $B^*$ may be subject to high variance. To tackle this problem, and to exploit possible structure in the parameterizations, we introduce a "generating" function $g$, which takes inputs $\hat{\beta} \in \mathbb{R}^p$ (with $p$ chosen a priori) and transforms them into outputs $\beta$. This function can be used to express both regularities and restrictions in the space of parameterization. For instance, in the case of a time varying model, the optimal, temporally local parameterizations of $f$ may lie on a low-dimensional manifold embedded in the full parameter space of $f$. The structure of such a manifold could be of interest, and restricting the estimation could significantly reduce variance in the resulting parameter estimates with only a small increase in bias.

We can now rephrase (1) as an optimization problem involving $g$. Given dimension $p$, a model $f$, a loss $\ell$, and a set of inputs $X$, our optimization becomes:

$$\arg\min_{g} \left[\min_{\hat{B}} \left[\ell(f|g, X, \hat{B})\right]\right] \qquad (5)$$

in which $\hat{B} = \{\hat{\beta}_1, ..., \hat{\beta}_{m'}\}$ is a set of inputs to $g$ and $f|g$ denotes the restriction of parameterizations of $f$ to the output space of $g$.

If we define $g(\hat{\beta}) \equiv \hat{\beta}$, then (5) exactly reproduces (1). If we allow $g$ to take an arbitrarily complex form, then we similarly recover the optimization in (1), as we can define $g(\hat{\beta}_i) \equiv \beta_i$ for each $\beta_i \in B^*$. Thus, interesting cases of (5) arise when $g$ is more carefully chosen. The next section illustrates some useful problems that arise from different definitions of $g$, $f$, and $\ell$.

## 3. Illustrations of the Problem Formulation

As a first example, consider performing a locally weighted regression analogous to that in (3), but with the local parameterizations of $f$ restricted to a linear subspace. Let $g(\hat{\beta}) = A\hat{\beta}^w$, where $A$ is the matrix of the parameters of $g$. We can re-write (5) as follows:

$$\arg\min_A \left[ \min_{\hat{B}} \left[ \sum_{i=1}^{m'} \sum_{j=1}^{m} k(\hat{\beta}_i^x, x_j)(x_j^T A \hat{\beta}_i^w - y_j)^2 \right] \right] \tag{6}$$

in which we now split each $\hat{\beta}_i$ into a localization subcomponent $\hat{\beta}_i^x$ and a coefficient subcomponent $\hat{\beta}_i^w$. If one views $x_j^T A \hat{\beta}_i^w$ as a reduction of $x_j$ into the subspace spanned by the columns of $A$, followed by a linear regression in that subspace, the objective in (6) is closely related to methods developed for linear dimension reduction for regression based on non-parametric estimators (Samarov, 1993; Xia et al., 2002). However, minor modifications, like regularizing the $\hat{\beta}_i$s via $\lambda \sum_i ||\hat{\beta}_i^w||_1$, weaken this link.

As a second example, we restate the mixture of Gaussians model under the constraint that the means $\{\beta_1^\mu, ..., \beta_{m'}^\mu\}$ of the parameterizations $\{\beta_1, ...\beta_{m'}\}$ lie within a linear subspace of the observation space, i.e. $\beta_i = (g(\hat{\beta}_i^\mu), \hat{\beta}_i^\Sigma, \hat{\beta}_i^\pi)$[1], with $g$ defined as for (6). The resulting optimization can be written as follows:

$$\arg\min_A \left[ \min_{\hat{B}} - \log \left( \prod_{j=1}^{m} \left[ \sum_{i=1}^{m'} \hat{\beta}_i^\pi p(x_j | A\hat{\beta}_i^\mu, \hat{\beta}_i^\Sigma) \right] \right) \right] \tag{7}$$

Performing the optimization in (7) was shown to be useful for classification tasks in (Sajama & Orlitsky, 2005).

We can similarly generate optimization problems in the form of (5) whose solutions correspond to PCA and sparse coding, which are left out due to space constraints.

## 4. Learning Compact Representations of Time Varying Network Structure

In this section, we use our new problem formulation to derive a novel algorithm for estimating time varying network structure, using a time-dependent sparse combination of learned basis structures. Through an analogy between our algorithm and sparse coding (Olshausen & Field, 1996), we then extend our algorithm to learning of task-driven basis structures, guided by the work in (Mairal et al., 2011). We begin by reviewing existing work on network structure estimation, before describing the new algorithms.

### 4.1. Sparse Network Structure Estimation

In recent years, much effort has gone into developing effective methods for estimating sparsely structured Gaussian graphical models. A Gaussian graphical model (GGM) explains a set of $m$ $n$-dimensional observations $X = \{x_1, ..., x_m\}, x_i \in \mathbb{R}^n$ using a set of $n$ vertices (each corresponding to one dimension) and a set of edges, each describing the strength of the relationship between its incident vertices. A GGM implies a covariance $\Sigma$ and is equivalent to modeling $X$ with a normal distribution $\mathcal{N}(\vec{0}, \Sigma)$. Typically, prior to estimating a GGM, the observations are standardized to have mean 0.

Many existing methods addressing GGMs focus on estimating their structure, i.e. the pattern of zero/non-zero edges. These methods typically work with the precision matrix (i.e. $\Sigma^{-1}$) implied by a GGM, as non-zero entries in $\Sigma^{-1}$ correspond to non-zero edges in the GGM. Estimating the structure of $\Sigma^{-1}$ is facilitated by the following relationship:

$$\rho_{ij} = \frac{\tilde{\sigma}_{ij}}{\sqrt{\tilde{\sigma}_{ii}\tilde{\sigma}_{jj}}}, \tag{8}$$

in which $\rho_{ij}$ indicates the partial correlation between the $i^{th}$ and $j^{th}$ dimension conditioned on the values of all other dimensions, and $\tilde{\sigma}_{ij}$ is the entry in the $i^{th}$ row and $j^{th}$ column of $\Sigma^{-1}$. The relationship between partial correlations and GGM structure leads to efficient methods for GGM structure estimation, as partial correlations can be directly estimated by "self-regression".

The use of self-regression for network structure estimation is based on the following results (Lauritzen, 1996):

$$x_t^i = \sum_{j \neq i} x_t^j \tilde{\rho}_{ij} + \epsilon_t^i, \tag{9}$$

in which $x_t^i$ represents the value of the $i^{th}$ dimension of the $t^{th}$ observation in $X$, $\tilde{\rho}_{ij}$ is a real-valued scalar, and $\epsilon_t^i$ is uncorrelated with $x_t^i$ if and only if:

$$\tilde{\rho}_{ij} = -\frac{\tilde{\sigma}_{ij}}{\tilde{\sigma}_{ii}} = \rho_{ij}\sqrt{\frac{\tilde{\sigma}_{jj}}{\tilde{\sigma}_{ii}}}, \text{ from which} \tag{10}$$

$$\rho_{ij} = \text{sign}(\tilde{\rho}_{ij})\sqrt{\tilde{\rho}_{ij}\tilde{\rho}_{ji}}. \tag{11}$$

Hence, $\tilde{\rho}_{ij}$ can be efficiently estimated for any given $i$ using linear regression of the response variables $\{x_1^i, ..., x_m^i\}$ on the covariates $\{x_1^{\setminus i}, ...x_m^{\setminus i}\}$, in which $x_t^{\setminus i}$ indicates a vector including all dimensions except $i$; $\rho_{ij}$ can then be computed as well.

Most existing methods for GGM structure estimation assume that $\Sigma^{-1}$ is sparse. Value estimation methods estimate each entry in $\Sigma^{-1}$ (Zhou et al., 2010), while structure estimation determines the pattern of zero/non-zero entries (Friedman et al., 2008; Song et al., 2009b; Kolar & Xing, 2011)). The sparsity assumption can be incorporated into the self-regression process by using sparsifying regression techniques, such as the well known Lasso (Tibshirani, 1996). Self-regression methods using sparsity have been shown to produce consistent estimates of structure in

[1] Note that we have *not* transformed the covariances $\hat{\beta}_i^\Sigma$

$\Sigma^{-1}$ under suitable conditions (Meinshausen & Bühlmann, 2006; Wainwright et al., 2007; Kolar & Xing, 2011).

A recent line of work focuses on extending methods for network structure estimation for the case when structures vary over time (Ahmed & Xing, 2009; Kolar et al., 2009; Song et al., 2009a;b; Zhou et al., 2010; Kolar & Xing, 2011). We focus on the KELLER algorithm from (Song et al., 2009a), as our algorithm can be seen as its natural generalization using the problem formulation in (5). The KELLER algorithm is predicated on two assumptions: sparsity in the time varying network structure, and smoothness in the changes of these structures over time. This second assumption distinguishes KELLER from methods such as (Ahmed & Xing, 2009) and (Kolar et al., 2009), which assume abrupt changes in the network structure.

To estimate the structure of a network at time $t$, given a sequence of $T$ observations $X = \{x_1, .., x_T | x_i \in \mathbb{R}^n\}$, KELLER performs a set of $n$ independent $\ell_1$-regularized locally weighted regressions, with the $i^{th}$ regression estimating the values $\tilde{\rho}_{ij}, \forall j \neq i$ as described above. By using locally weighted regression, these values are specifically adapted to the predominant network structure affecting the observation at time $t$. For time $t$, these regressions can be written compactly as follows:

$$A_t^* = \underset{A \in \mathbb{R}^{n \times n}}{\arg\min} \sum_{t'=1}^{T} k(t, t') \, ||x_{t'} - A x_{t'}||_2^2 + \lambda ||A||_1, \quad (12)$$

in which $k(t, t')$ computes a kernel weight measuring temporal proximity, diagonal entries of $A$ are fixed at 0, $||A||_1$ is the entry-wise matrix 1-norm (i.e. $\sum_i \sum_j |A_{ij}|$), and $\lambda$ controls the $\ell_1$ regularization, which determines the sparsity of $A$. After estimating $A^*$ according to (12), KELLER performs a simple procedure to make the implied structure estimate coherent with the assumption of an undirected GGM (i.e. $A^*$ should be symmetric), which consists of inferring an edge between any pair of vertices $(i, j)$ such that $A_{ij} \neq 0$ or $A_{ji} \neq 0$. An estimation similar to (12) is used in (Song et al., 2009b), without the additional symmetrization, for networks with directed edges.

As used in (12), the weighting kernel makes the estimate of $A_t^*$ at time $t$ effectively independent from observations at times remote from $t$. This can lead to high variance, and ignores potential structure in the way in which the network structure changes over time. We will now state our algorithm, which addresses these problems.

### 4.2. Estimating Network Structures as Combinations of Basis Structures

We reformulate the optimization in (12) similarly to the way in which we generalized locally weighted regression from (3) to the form (6). At each time $t$, the optimal $A_t^*$ is estimated as a linear combination of a set of $k$ basis matrices $\hat{A} = \{A^1, ..., A^k | \text{ diag}(A^i) = 0\}$. Our proposed estimation procedure revolves around the following optimization:

$$\hat{\beta}_t = \underset{\hat{\beta} \in \mathbb{R}^k}{\arg\min} \sum_{t'=1}^{T} k(t, t') \, ||x_{t'} - \sum_{i=1}^{k} \hat{\beta}^i A^i x_{t'}||_2^2 + \lambda \, r(\hat{\beta}) \quad (13)$$

in which $\hat{\beta}^i$ is the $i^{th}$ element of $\hat{\beta}$, $r(\hat{\beta})$ is a regularization term, and $\lambda$ controls the strength of regularization. Given $\hat{\beta}_t$, we estimate $A_t^*$ as $\sum_{i=1}^k \hat{\beta}_t^i A^i$. The optimization in (13) involves a fixed set of basis matrices $\hat{A}$, but what we really want is to jointly optimize the loss in (13) over all times $1 \leq t \leq T$, with respect to both the $\hat{\beta}_t \in \hat{B} = \{\hat{\beta}_1, ...\hat{\beta}_T\}$ and the $A^i \in \hat{A}$. By doing so, information extracted from the entire sequence is allowed to affect the estimation of each $A_t^*$ when each $A_t^*$ is constructed from the bases in $\hat{A}$, which helps mitigate problems with high variance.

The desired joint optimization over $\hat{B}$ and $\hat{A}$ is easy to express in the terms of (5). Let $g(\hat{\beta}) = \sum_{i=1}^k \hat{\beta}^i A^i$, where $A^i \in \hat{A}$ and $||A^i||_1 \leq c$. The constraint on the entry-wise 1-norm of each $A^i$ enforces the structural sparsity assumption. Next, we define $f(x, g(\hat{\beta})) = ||x - g(\hat{\beta})x||_2$. Then, we define $\ell(f|g, X, \hat{B})$ as:

$$\ell(f|g, X, \hat{B}) = \sum_{t=1}^{T} \sum_{t'=1}^{T} k(t, t') \, f(x_{t'}, g(\hat{\beta}_t))^2 + \lambda \sum_{t=1}^{T} r(\hat{\beta}_t) \quad (14)$$

Finally, we express the full joint optimization as follows:

$$\hat{A}^* = \underset{\hat{A}}{\arg\min} \min_{\hat{B}} \sum_{t=1}^{T} \sum_{t'=1}^{T} k(t, t') \, ||x_{t'} - \sum_{i=1}^{k} \hat{\beta}_t^i A^i x_{t'}||_2^2 + \lambda_\beta \sum_{t=1}^{T} r(\hat{\beta}_t) + \lambda_A \sum_{i=1}^{k} ||A^i||_1 \quad (15)$$

in which we changed the entry-wise 1-norm constraint on each $A^i$ for a functionally similar entry-wise 1-norm regularization term. Intuitively, our method produces a set of basis network structures, i.e. $\hat{A}^*$, with which the temporally local network structures can be effectively approximated.

The joint optimization in (15) is closely analogous to the following sparse coding objective:

$$A^* = \underset{A \in \mathbb{R}^{n \times k}}{\arg\min} \left[ \min_{B} \sum_{i=1}^{m} \left( ||x_i - A\beta_i||_2^2 + \lambda ||\beta_i||_1 \right) \right], \quad (16)$$

in which $B = \{\beta_1, ...\beta_m \, | \beta_i \in \mathbb{R}^k\}$, $\lambda$ controls the tradeoff between reconstruction accuracy and representational sparsity, and the columns of $A$ are constrained to unit norm. We can emphasize this by introducing the concept of time varying *pseudo-dictionaries* $D_t \in \mathbb{R}^{n \times k}$, in which the $i^{th}$ column of $D_t$ is $A^i x_t$. Using pseudo-dictionaries, we can rewrite (15) as follows:

$$\hat{A}^* = \arg\min_{\hat{A}} \tag{17}$$
$$\min_{\hat{B}} \sum_{t=1}^{T} \left[ \left( \sum_{t'=1}^{T} k(t,t') \, ||x_{t'} - D_{t'}\hat{\beta}_t||_2^2 \right) + \lambda_\beta r(\hat{\beta}_t) \right]$$

in which we dropped the sparsifying penalty on $A^i \in \hat{A}^*$ for notational brevity. From (17), it can be seen that the inner optimization over $\hat{B}$ in (15) can be addressed as a set of sparse coding problems. For our purposes, we set the regularization term $\lambda_\beta r(\hat{\beta}_t)$ to:

$$\frac{\alpha\lambda_\beta}{2}||\hat{\beta}_t||_2^2 + (1-\alpha)\lambda_\beta||\hat{\beta}_t||_1; \; 0 \le \alpha \le 1, \tag{18}$$

which corresponds to elastic-net regularization (Zou & Hastie, 2005). We use this form to meet the assumptions required for the task-driven dictionary learning described in (Mairal et al., 2011), used in the further extension of our algorithm.

The analogy between our method and sparse coding leads naturally to a method for effecting the joint optimization in (15). As in sparse coding, we can jointly optimize over $\hat{A}$ and $\hat{B}$ using an EM-like block coordinate descent process that alternates between optimizing $\hat{B}$ while holding $\hat{A}$ fixed and optimizing $\hat{A}$ while holding $\hat{B}$ fixed (each of these is a convex problem). When optimizing $\hat{B}$ with $\hat{A}$ held fixed, we compute the optimal $\hat{\beta}_t$ for each $t$ via elastic-net regressions solved with the publicly available, highly optimized glmnet package (Friedman et al., 2009). When optimizing $\hat{A}$ with $\hat{B}$ held fixed, given current estimates of each basis $A^i \in \hat{A}$, we compute the partial gradients of the objective in (17) w.r.t. the entries of each pseudo-dictionary $D_t$, and then backpropagate these partial gradients through the pseudo-dictionary formation process to get partial gradients w.r.t. each entry of each basis structure $A^i$. We symmetrize the partial gradient of (17) w.r.t. each $A^i$ by setting $\partial A^i_{uv} = \frac{1}{2}(\partial A^i_{uv} + \partial A^i_{vu})$. We also set $\partial A^i_{uu} = 0, \forall u$ to maintain the zero-diagonal constraint on $A^i \in \hat{A}$. In the next subsection we refer to these (unsupervised) partial gradients as $\nabla_{A^i}\ell_u$. Using the computed gradients, we then take a single gradient descent step to update each $A^i$.

The full joint optimization process iterates between updating the $\hat{\beta}_i \in \hat{B}$ via the regression in (13) and performing a single gradient descent update of the entries in each $A^i \in \hat{A}$. We dynamically select the step size for gradient descent updates in each iteration by line search and iterate until convergence. We perform the iterative optimization using subsampled batches of the available observations, which yields a stochastic gradient descent approach to jointly optimizing (15)/(17).

### 4.3. Supervised Basis Structure Learning

We can adapt the work of (Mairal et al., 2011) to enable our algorithm to learn task-driven sets of basis network structures. We consider the task of minimizing differentiable supervised loss functions that can be written as:

$$L_s(X, \hat{B}, w) = \sum_{t=1}^{T} \ell_s(\omega^\top \hat{\beta}_t, y_t) + \frac{\nu}{2}||\omega||_2^2, \tag{19}$$

where $\omega \in \mathbb{R}^k$, $y_t$ is the target output at time $t$, and the $\hat{\beta}_t \in \hat{B}$ were produced to minimize (17). This includes any differentiable linear function of the $\hat{\beta}_t \in \hat{B}$. In this paper, we focus on classification tasks and thus use the binomial deviance loss of logistic regression, i.e. $\ell_s(\omega^\top\hat{\beta}_t, y_t) = \log(1 + e^{-y_t\omega^\top\hat{\beta}_t}),\; y_t \in \{-1, +1\}$.

The crux of task-driven dictionary learning is converting the readily available gradients of $\ell_s$ w.r.t. the structure codes $\hat{\beta}_t$ into gradients w.r.t. the pseudo-dictionaries $D_t$ with which they were computed to minimize (17), as gradients w.r.t. the $D_t$ easily produce gradients w.r.t. the $A_i \in \hat{A}$. Unfortunately, the optimization producing the $\hat{\beta}_t$ makes the conversion $\nabla_{\hat{\beta}_t} \to \nabla_{D_t}$ non-trivial. However, Mairal et al. (2011) show that if elastic-net regularization is used to produce each $\hat{\beta}_t$ from the $(x_t, D_t)$, the gradient of the per-instance supervised loss $\ell_s$ w.r.t. $D_t$ can be computed as follows:

$$\nabla_{D_t}\ell_s(\omega^T\hat{\beta}_t, y_t) = -D_t\phi_t\hat{\beta}_t^\top + (x_t - D_t\hat{\beta}_t)\phi_t^\top, \tag{20}$$

in which $\phi_t \in \mathbb{R}^k$ is defined as follows:

$$\phi_{t\Lambda^C} = 0, \;\; \phi_{t\Lambda} = (D_{t\Lambda}^\top D_{t\Lambda} + \alpha\lambda_\beta I)^{-1}\nabla_{\hat{\beta}_{t\Lambda}}\ell_s(\omega^\top\hat{\beta}_t, y_t) \tag{21}$$

where $\Lambda$ denotes the indices of non-zero entries in the sparse $\hat{\beta}_t$, $\Lambda^C$ indicates the complementary set of indices, and $\alpha\lambda_\beta$ is the $\ell_2$ regularization weight from (18). Once gradients of $\ell_s$ w.r.t. each $D_t$ (i.e. $\nabla_{D_t}\ell_s$) have been computed for each time $t$, they can be backpropagated through the pseudo-dictionary formation process and summed across time points to get gradients w.r.t. each $A^i \in \hat{A}$ (i.e. $\nabla_{A^i}\ell_s$).

Given unsupervised gradients $\nabla_{A^i}\ell_u$, computed as described at the end of Sec. 4.2, and supervised gradients $\nabla_{A^i}\ell_s$, we define the final gradients for stochastic descent optimization of the combined unsupervised/supervised objective as follows:

$$\partial A^i = \gamma\nabla_{A^i}\ell_u + (1-\gamma)\nabla_{A^i}\ell_s; \;\; 0 \le \gamma \le 1, \tag{22}$$

where $\gamma$ is a mixing parameter controlling the tradeoff between supervised and unsupervised learning. As before, we enforce symmetry and zero-diagonal constraints prior to using the joint gradients for basis updates.

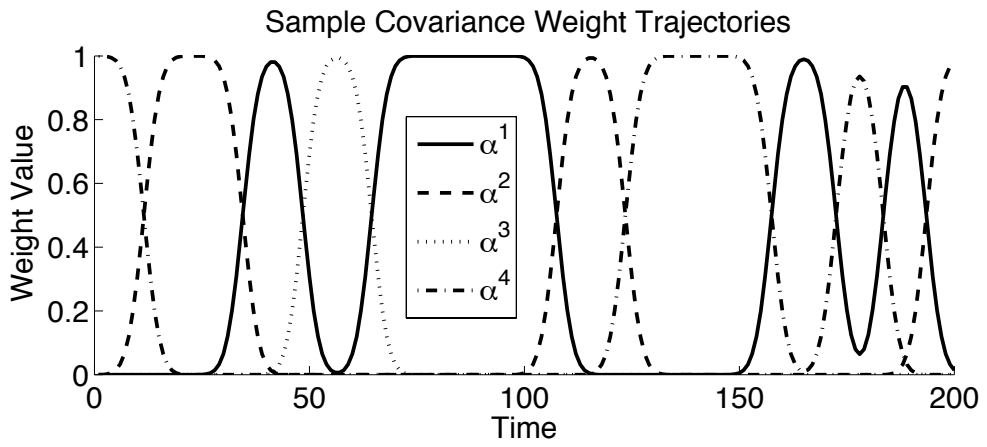

*Figure 1.* Trajectories for the weights $\alpha_t^i$ used in the tests described in Sec. 4. Note the smooth transitions between "structural regimes", which cause problems for methods expecting abrupt structural changes.

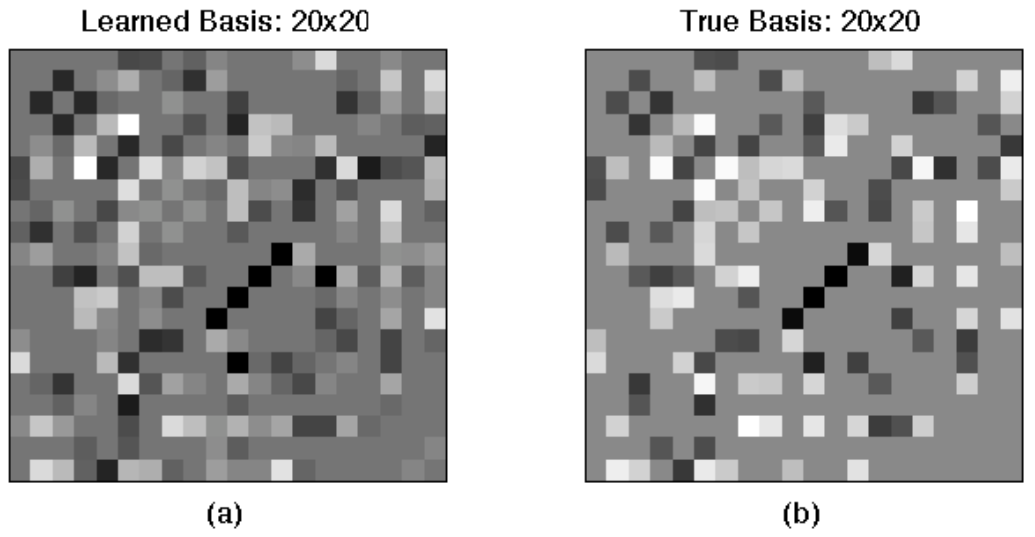

*Figure 2.* The left panel is the best-match learned basis for the true basis on the right (taken from one of the test sequences). This pairing represents a match quality within one standard deviation of the mean for this network size. Gross structural similarities are readily apparent. The diagonal of the true basis has been removed to facilitate comparison.

# 5. Synthetic Network Analysis

This section presents tests based on simulated observation sequences which show the ability of our algorithm to recover recurring elements of time varying network structures. We generated each observation sequence by drawing the observation $x_t$ at time $t$ from a normal distribution $\mathcal{N}(0, \Sigma_t)$, in which $\Sigma_t$ was a convex combination of four covariance matrix bases: $\Sigma_t = \sum_{i=1}^4 \alpha_t^i \Sigma^i$, with $\sum_{i=1}^4 \alpha_t^i = 1$ and $0 \leq \alpha_t^i \leq 1$. We generated smooth trajectories for the $\alpha_t^i$ (an example set of trajectories is shown in Fig. 1). We generated each $\Sigma^i$ by symmetrically removing two thirds of the off-diagonal entries (the ones with the smallest magnitude) from a random covariance matrix with eigenvalues uniformly distributed in $(0, 1)$, and then rescaling diagonal entries to ensure positive definiteness. An example of the sparse basis structures used in our tests can be seen in the right panel of Fig. 2. The inputs ranged from 10-dimensional to 40-dimensional. For each tested dimensionality, we generated 25 sequences of 5000 observations, with the first 3000 reserved for training and the last 2000 reserved for testing; each sequence was based on different basis matrices $\Sigma^i$ and different $\alpha_t^i$ trajectories. Results are averaged over the 25 sequences.

Methods based on (12) are much better suited for this task than methods expecting abrupt "change point" structure. Hence, we tested three methods for estimating time varying network structure in our sequences: locally weighted $\ell_1$-regularized self-regression (as described in (12)), the same self-regression followed by projection of the inferred structures onto the principal components of structures estimated for each time point in the training set, and our iterative approach to learning task-driven basis structures.

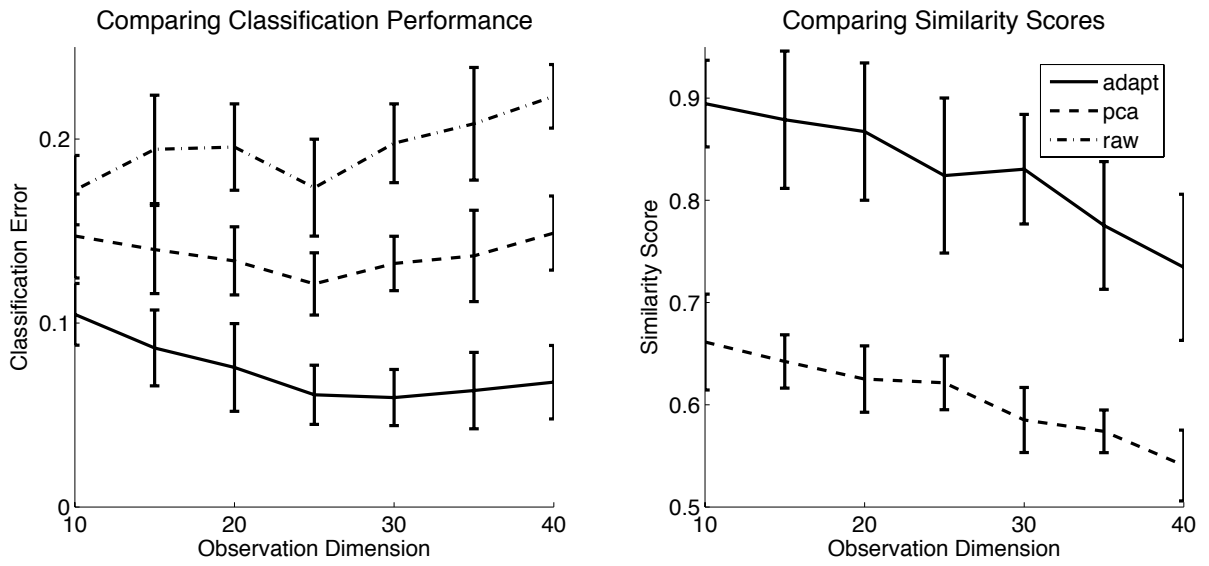

*Figure 3.* Left plot: mean classification error for a binary classification task as a function of input dimension. Our adaptive basis learning method (label: adapt) generates better features than the self-regression-based (label: raw) and PCA-based (label: pca) approaches. Right: the fidelity of the principal structures produced by using PCA on the $A_t^*$ computed according to (12) and the basis structures produced by adapting these principal structures to minimize (15). The bases produced by our method capture more accurately the generative structure underlying the test sequences

The self-regression-based method used in our tests can be considered equivalent to KELLER (Song et al., 2009a). Using the principal components of the set of $A_t^*$ produced by this method is itself novel, and can be seen as an approximation to our method. When executing our method, we initialized the set $\hat{A}$ using these principal structures. In our tests, we used six principal structures with the PCA-based method and learned six basis structures with our algorithm.

We measured test performance for a classification task in which the class of each $x_t$ was set as follows: $y_t = 1$ if $\alpha_t^1 + \alpha_t^2 \geq \alpha_t^3 + \alpha_t^4$ and $y_t = -1$ otherwise. We also estimated a similarity score between the sets of estimated structures and the true precision matrices underlying each sequence, as explained below.

Classification was performed using the parameterization produced by each method for a given $x_t$ (i.e. a matrix $A_t^*$ for the self-regression method, the same matrix projected onto a set of principal structures for the PCA method, and the inferred vector $\hat{\beta}_t$ for our algorithm) as input features to a regularized logistic regression classifier, with the target class determined by $y_t$. Fig. 3 presents the results. The basis structures learned by our method, and the codes they induce, offer an informative representation of regularities in time varying sparse network structure.

We measured similarity between learned bases and the true precision matrices using a form of pairwise matrix correlation. First we set the diagonal entries of each matrix to zero, then their off-diagonal entries to zero mean and unit norm, and finally "vectorize" each matrix and compute the dot product between the resulting vectors. This measure ranges from $-1$ to 1, with larger magnitudes indicating greater similarity. For each sequence and each method, we found the best match to each $(\Sigma^i)^{-1}$, as determined by the magnitude of our correlation score, among the set of bases produced by that method. We then averaged best match scores for each method over both true bases and sequences, to get a final score for each dimensionality. Fig. 3 shows the similarity scores achieved by the PCA-based method and our method, with the bases produced by our method consistently displaying greater similarity to the true bases than those produced by PCA alone. Fig. 2 shows a typical example of a best match produced by our method during these tests; as can be seen, the learned basis is qualitatively very similar to the true basis.

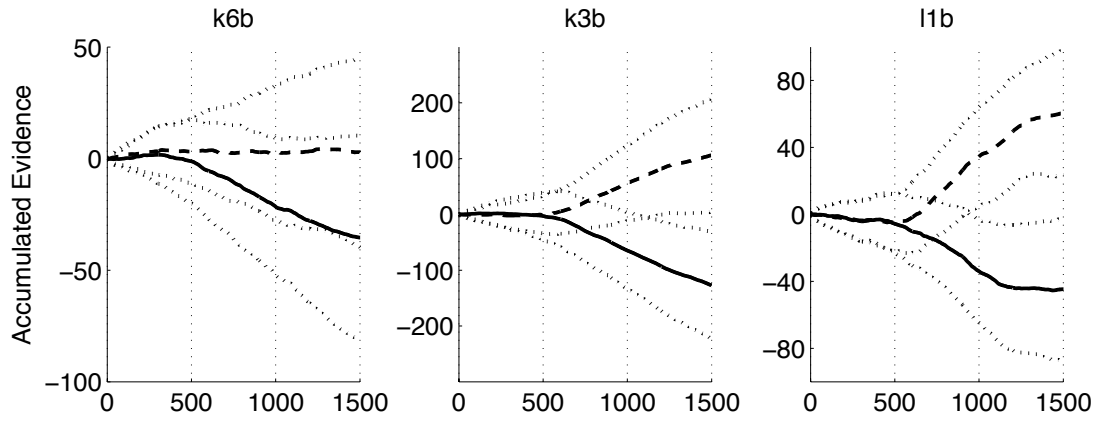

*Figure 4.* Behavior of the evidence accumulation classifier learned using features produced by our algorithm, averaged over all trials for each subject. As indicated by Table 1, subject k6b proved difficult for both our method and RCSP. With the other two subjects, the discriminative capacity of our bases can be clearly seen in the rapid bifurcation of their induced classifier response after the cue time at $x = 500$. Lines trending upwards indicate left hand trials and lines trending downwards represent right hand trials.

## 6. BCI EEG Analysis

We applied our algorithm to the analysis of EEG data from a Brain Computer Interface (abbr. BCI) motor imagery experiment available as task 3a from BCI competition III (Schlögl et al., 2005; Blankertz et al., 2006). In this task, the objective is to infer the motor action visualized by a subject during a set of test trials, given a labeled set of training trials. In each trial, a cue is given to the subject indicating a motor action, after which the subject visualizes that action for several seconds. Cortical activity during each trial was measured by a set of 60 electrodes placed on the scalp, taking measurements at 250Hz. Data collected from these electrodes was the subject of our analysis.

We used left hand and right hand trials from this dataset for the subjects l1b, k3b, and k6b. Several trials from each subject were discarded due to significant artifacts, as measured by deviation from a Gaussian model of the mean behavior of the joint set of trials for a subject. We also applied a whitening transform $VD^{-\frac{1}{2}}V^T$ to each subject's data prior to analysis, where $D$ was a diagonal matrix containing the eigenvalues of the data and the columns of $V$ were the corresponding eigenvectors. We set kernel widths and regularization weights for the optimization in (15) uniformly for all subjects and trials, following a brief manual search.

| | L1b | K6b | K3b |
|---|---|---|---|
| RCSP | 0.100 (0.056) | 0.363 (0.118) | 0.103 (0.048) |
| ADAPT | 0.041 (0.053) | 0.330 (0.098) | 0.056 (0.033) |
| JOINT | 0.052 (0.052) | 0.253 (0.080) | 0.063 (0.039) |

*Table 1.* Classification error means and standard deviations for each subject in a set of BCI motor imagery experiments for Regularized Common Spatial Patterns (RCSP), our algorithm (ADAPT), and a classifier based on the combined set of features from the previous two (JOINT). Our algorithm outperforms a method used in current practice, with a classifier built on the joint feature set producing the best performance when averaged across all subjects.

We learned a set of 20 sparse basis structures for each subject using our algorithm in an unsupervised fashion (i.e. $\gamma = 1$). Afterwards, we performed 20 rounds of randomized cross-validation in which we split the trials for each subject 4/1 into training/test sets. We trained three classifiers in each round of cross-validation: a classifier built on the $\hat{\beta}_t$ inferred by our algorithm after a period of supervised basis updates (i.e. $\gamma = 0.75$) using the training set, a classifier built on the output of a set of 20 RCSP filters (Lotte & Guan, 2011), and a classifier built on the combination of both feature sets. The regularization parameter for RCSP was selected to maximize expected performance across all subjects.

We built our classifier by considering the $\hat{\beta}_t$ and class labels for each time point in each training trial as inferred feature/label pairs for training an $\ell_2$-regularized logistic regression classifier. Given the encoding of a particular trial in terms of a set of $\hat{\beta}_t$, an overall output for the trial was computed by accumulating (i.e. summing) the output of the learned single time-point classifier over the first three post-cue seconds of the trial. After this evidence accumulation phase, the classification for each trial was determined by the sign of its overall output. RCSP filters were trained as described in (Lotte & Guan, 2011), after which the squared responses of these filters to the observations were used as input features to an $\ell_2$-regularized logistic regression classifier, trained as for our algorithm. We also trained an analogous classifier using the combined features produced by our algorithm and the RCSP filters at each time point. Classification results for each subject are shown in Table 1, and a visual representation of the evidence accumulation process based on our features is shown in Fig. 4. Classifiers constructed in this fashion have the advantage of being amenable to "early exit", in the spirit of drift-diffusion decision making.

These results show that our approach produces informative features in a real-world scenario, with the results for the combined features suggesting that our features supplement, rather than replace, the commonly used RCSP features.

## 7. Conclusion

We introduced a problem formulation in the context of multiply parameterized models. Using this formulation, we developed a novel algorithm for learning representations of sparse structure in time varying networks with recurring structural motifs. We used tests on synthetic data to show that our algorithm behaves as desired under suitable conditions, while an application to BCI EEG data showed the potential value of our algorithm in real world conditions.

We plan to apply our approach to other types of tasks, such as analysis of time varying weather and traffic patterns, in addition to investigating alternative parameter transformation methods, beyond the linear transforms considered in this paper. Our algorithm is readily extensible to the estimation of time varying structure in Dynamic Bayesian Networks.

**Acknowledgements**: Funded by NSERC and ONR.